\documentclass[journal]{IEEEtran}

\ifCLASSINFOpdf
\else
   \usepackage[dvips]{graphicx}
\fi
\usepackage{url}

\pdfoutput=1

\usepackage{amsmath}
\usepackage{indentfirst}
\usepackage{amssymb}
\usepackage{subfigure}
\usepackage{algorithm}
\usepackage{algorithmic}
\usepackage{graphicx}
\usepackage{multirow}
\usepackage{hyperref}
\usepackage{times}
\usepackage{epsfig}
\usepackage{subfigure}
\usepackage[justification=centering]{caption}
\usepackage{graphicx,float}
\usepackage{amsmath}
\usepackage{amssymb}
\usepackage{multirow}
\usepackage{booktabs}       % professional-quality tables
\usepackage{epstopdf}
\usepackage{verbatim}
\begin{document}
	
\title{Branch-Cooperative OSNet for Person Re-Identification}
\author{Lei Zhang, Xiaofu Wu$^\dag$, Suofei Zhang and Zirui Yin% <-this % stops a space
%\author{Lei Zhang}
\thanks{$^\dag$Corresponding author. This work was supported in part by the National Natural Science Foundation of China under Grants 61372123, 61671253 and by the Scientific Research Foundation of Nanjing University of Posts and Telecommunications under Grant NY213002.}% <-this % stops a space
\thanks{Lei~Zhang, Xiaofu~Wu and Zirui~Yin are with the National Engineering Research Center of Communications and Networking, Nanjing University of Posts and Telecommunications, Nanjing 210003, China (E-mails: 1019010621@njupt.edu.cn; xfuwu@ieee.org; 1219012816@njupt.edu.cn).}
\thanks{Suofei Zhang is with the School of Internet of Things, Nanjing University of Posts and Telecommunications, Nanjing 210003, China (E-mail: zhangsuofei@njupt.edu.cn).}}

\maketitle
\begin{abstract}
Multi-branch is extensively studied for learning rich feature representation for person re-identification (Re-ID). In this paper, we propose a branch-cooperative architecture over OSNet, termed BC-OSNet, for person Re-ID. By stacking four cooperative branches, namely, a global branch, a local branch, a relational branch and a contrastive branch, we obtain powerful feature representation for person Re-ID. Extensive  experiments show that the proposed BC-OSNet achieves state-of-art performance on the three popular datasets, including Market-1501, DukeMTMC-reID and CUHK03. In particular, it achieves mAP of 84.0\% and rank-1 accuracy of 87.1\% on the CUHK03\_labeled.
\end{abstract}


\IEEEpeerreviewmaketitle

\newtheorem{plm}{Problem}
\newtheorem{thm}{Theorem}
\section{Introduction}
The increasing demand makes deep learning methods account for a large proportion in the field of computer vision, such as image classification, target detection, semantic segmentation, person re-identification(Re-ID), etc. Often, there are three typical functions for a person Re-ID system: person detection, person tracking and person retrieval. In general, the task of person Re-ID focuses on person retrieval, which is to train a feature descriptor with person-discriminative capabilities, from large-scale pedestrians captured by multiple cameras. The main difficulty for person Re-ID comes from the fact that pedestrian images captured by cameras with a range of differences including perspectives, image resolutions, changes in illumination, unconstrained posture, occlusion, heterogeneous modes\cite{Zheng2016}. It still remains a big challenge for further improving the retrieval accuracy.

The key step for person Re-ID system is to find a rich but discriminative feature representation for pedestrian images. In the past decade, convolutional neural networks (CNNs) have been widely used in Re-ID tasks\cite{Krizhevsky2012} due to their attractive advantages. Usually, CNNs can only use global features for pedestrian images. However, the retrieval performance is very limited because intra-class variations cannot be well represented by global features. In order to eliminate this limitation, many methods have been proposed. Pose-based Re-ID \cite{Su2017PDC} proposed to use the key points of pose to divide the body and weight different blocks for enhancing the feature representation for recognition. Part-based Re-ID (PCB, MGN)\cite{Sun2018PCB}\cite{Wang2018MGN} were proposed for local feature representation (such as head, body, etc.) without using pose estimation. In order to take into account the connection between parts of images and contrast of background and pedestrian, relation module and global contrastive pooling(GCP)\cite{Park2019GCP} were proposed. Considering the lightweight of CNN architecture, OSNet\cite{Zhou2019OSNet} performs very well and is capable of learning omni-scale feature representations. 

In this paper, we propose a branch-cooperative OSNet for person Re-ID. By combining various branch-oriented features, including part-level and global-level features, relational and contrastive features,  BC-OSNet captures more feature details for retrieval. In brief, the main contributions of this paper are as follows:
We summarise our contribution as follows:
\begin{enumerate}

\item
Based on the baseline OSNet \cite{Xie2020PLR-OSNet}, we propose a branch-cooperative network architecture for person Re-ID. The proposed BC-OSNet has four branches, including a global branch, a local branch with part-level features, the relational branch and contrastive branch. We show that these branches are cooperative for enriching the feature representation.

\item
Extensive experiments show that the proposed BC-OSNet can achieve state-of-the-art results on the popular Re-ID datasets, despite of its small size. In particular, our results on CUHK03\_labeled might be the best, achieving mAP of 84.0\% and rank-1 accuracy of 87.1\% on the CUHK03\_labeled.
\end{enumerate}

\section{Related Work}

\subsection{Part-Level Features}
In general, global features are helpful for learning the contour information, so that the images can be retrieved from a broader perspective. Part-level features, however, may contain more fine-grained information. In \cite{Yi2014}\cite{Li2014}, the input pedestrian image was divided into three overlapping parts, and then the three part-level features can be well learned. Later, different ways of body dividing appeared. Pose-driven Deep Convolutional (PDC)\cite{Su2017PDC} took into account the posture of the body into the influence of appearance and employed the pose estimation algorithm to predict the posture. Part-Aligned Representations (PAR)\cite{Zhao2017PAR} cut the human body into several distinguishing regions and then connects feature vectors from each region in obtaining the final feature vector representation. Part-based Convolution Baseline (PCB)\cite{Sun2018PCB} learns the part-level features by dividing the feature map equally, and Refined Part Pooling (RPP) was proposed to improve the content consistency of the divided area. For Multiple Granularities Network (MGN)\cite{Wang2018MGN}, it uniformly divides the image into multiple stripes for obtaining  a local feature representation with multiple granularities.

\subsection{Relational and Contrastive Features}
The basic idea behind the relation network is to consider all entity pairs and integrate all these relations\cite{Santoro2017}.The concatenation of both global-level and part-level features is beneficial for rich personnel representation. To further improve the feature richness, the relation between part-level and the rest, and the contrastive information between background and retrieved object are equally important. This will help to  build a link between various parts since they often do not function independently. Global contrastive pooling(GCP)\cite{Park2019GCP} aggregates most discriminative information while One-vs.-rest relation module\cite{Park2019GCP} utilizes the relation between part-level and the rest to make the network more distinguishable, while retaining the compact feature representation for person Re-ID.

\subsection{Other Related Works}
With max-pooling as downsampling for CNNs, the relative position information for features is of more importance. In order to reduce the model size, global average pooling was proposed to replace the fully-connected layer with better overfitting performance. Recently, generalized-mean (GeM)\cite{Radenović2018GeM} pooling was proposed for narrowing the gap between max-pooling and average-pooling. Due to the success of the dropout\cite{Hinton2012dropout}, several variants, such as fast dropout\cite{Wang2013fastdropout} and DropConnect\cite{Wan2013dropconnect} were proposed. A continuous dropout algorithm\cite{Shen2017GCDropout} was proposed for achieving a good balance between the diversity and independence of subnetworks. In this paper,  we propose to employ batch dropblock\cite{Dai2019BDB} in our architecture.  Unlike the general dropblock\cite{Ghiasi2018dropblock}, Batch DropBlock\cite{Dai2019BDB} is an attentive feature learning module for metric learning tasks, which randomly drops the same region of all the feature maps in a batch during training and reinforces the attentive feature learning of the remaining parts.

\begin{figure*}
	\begin{center}
		\includegraphics[width=0.95\textwidth]{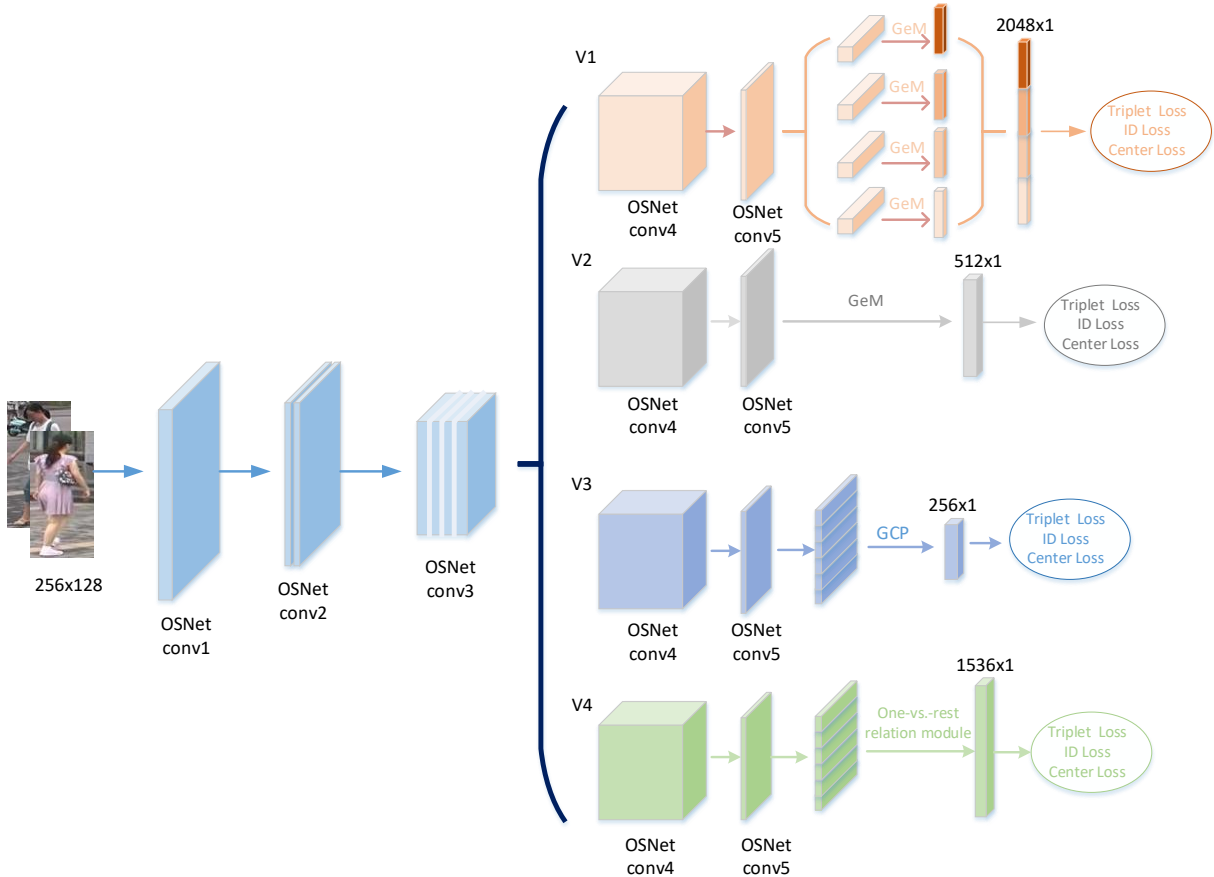}
	\end{center}
	\caption{Overall network architecture of BC-OSNet, four feature vectors from branches are concatenated for final testing}
	\label{fig:architecture}
\end{figure*}

\section{BC-OSNet}
 The overall network architecture of BC-OSNet is shown in Figure \ref{fig:architecture}, where the input is of size $ H \times W \times C $ with $ H, W, C(=3)$ denoting height, width and the number of channels, respectively. The shared-net of BC-OSNet takes the first 5 layers of  OSNet \cite{Zhou2019OSNet}, including 3 convolutional layers and 2 transition layers. Then, four cooperative branches are employed for feature extraction, including local branch (v1), global branch (v2), global contrastive pooling (GCP) branch (v3) and one-vs-rest relation branch (v4) \cite{Park2019GCP}, respectively. The use of four branches is to facilitate the learning of diverse but discriminative features.

\subsection{Cooperative Branches}
\subsubsection{Local Branch}
The first branch (v1) is a local branch. In this branch, the feature map is divided into 4 horizontal grids, and part-level features of $ 1 \times 1 \times C $ can be finally obtained by the use of  average pooling(AP). It should be noted that 4 part-level features are concatenated into a column vector for producing a single ID-prediction loss, while each part-level feature is driven by a ID-prediction loss for PCB \cite{Sun2018PCB}. Let
\begin{eqnarray}
\mathbf{f} = [f_1^T,f_2^T,\cdots,f_4^T]^T
\end{eqnarray}
denote the concatenated feature vector,  where $ f_1, f_2, f_3 f_4 $ denote 4 column vectors that divide the feature map horizontally.
Let the labeled set denote by $ \left\lbrace (x_i, y_i), i = 1,2, \cdots, N_s \right\rbrace $. Then, the ID-prediction loss can be written as
\begin{eqnarray}
\label{E2}
L=-\frac{1}{N_{s}} \sum_{i=1}^{N_{s}} \log \left(\frac{\exp \left(\left(\mathbf{W}^{y_{i}}\right)^{T} \mathbf{f}^{i}+b_{y_{i}}\right)}{\sum_{j} \exp \left(\left(\mathbf{W}^{j}\right)^{T} \mathbf{f}^{j}+b_{j}\right)}\right)
\end{eqnarray}
where $ W^{y_i} $ and $ W^j $ are the $ y_i-th $ and $ j-th $ columns of the weight matrix $ W $. Compared with PCB, this can obtain more effective and more differentiated information.
Usually, GAP is used for obtaining each part-level feature vector. Currently, both GAP and GMP were employed in existing methods and it is not well understood which pooling method is better. Here, we employ GeM \cite{Radenović2018GeM}, which 
\begin{eqnarray}
%\begin{split}
GeM(f_k = [f_0,f_1,\cdots,f_n])=[\frac{1}{n}\sum_{i=1}^{n}(f_i)^{p_k}]^\frac{1}{p_k}
%\end{split}
\end{eqnarray}
where GeM operator $f_k$ is a single feature map, when $ p_k \to \infty $, GeM is equal to max pooling, when $ p_k = 1 $, GeM is equal to average pooling, We initialized the GeM parameter with $ p_k = 1 $ in the local branch.

\begin{figure*}
	\begin{center}
		\includegraphics[width=0.9\textwidth]{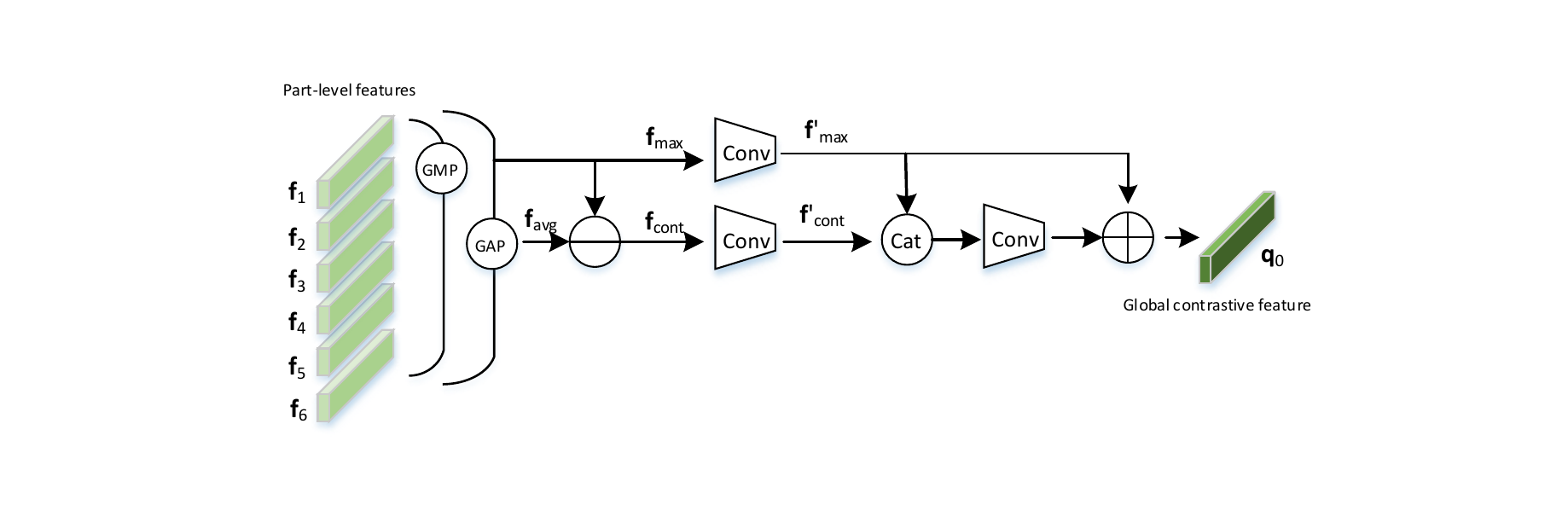}
	\end{center}
	
	\caption{global contrastive pooling(GCP)}
	\label{fig:gcp}
\end{figure*}

\begin{figure*}
	\begin{center}
		\includegraphics[width=0.8\textwidth]{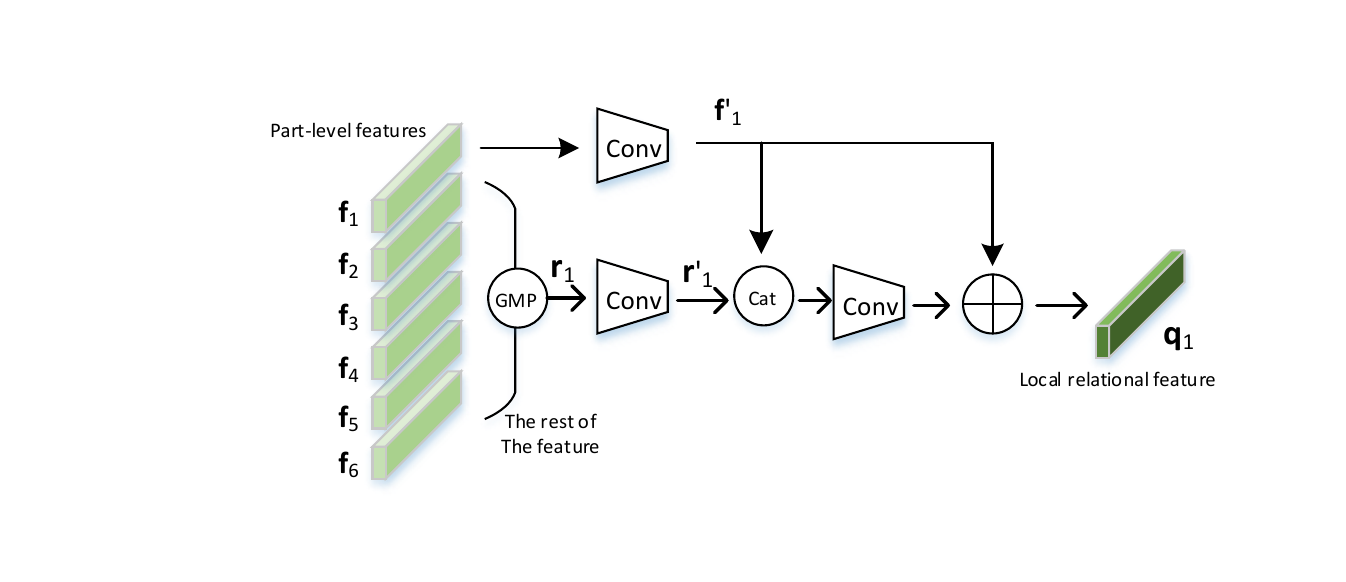}
	\end{center}
	\caption{One-vs.-rest relation module for computing local relational feature $\mathbf{q}_1$, $\mathbf{q}_2 \sim \mathbf{q}_5$ are not shown but computed in a similar way}
	\label{fig:ovsr}
\end{figure*}

\subsubsection{Global Branch}
The second branch (v2) is the global branch. The difference from the local branch is that GeM pooling is performed directly after conv4 and conv5. Note that we set $ p_k = 6.5 $ for  the initialization of GeM, and a 512-dimensional vector is obtained.  

\subsubsection{GCP Branch}
The third branch (v3) is a global contrastive pooling (GCP) branch. GCP is based on GAP and GMP to obtain local contrastive features. After obtaining the feature map at the output of conv5, it is divided into 6 horizontal grids and 256-dimensional feature vector is obtained by GCP. For better understanding the internal structure of GCP, we denote by $\mathbf{f}_{avg}$ and $\mathbf{f}_{max}$ the feature maps obtained with average pooling and max pooling, respectively. Note here that average pooling is operated on each part-level feature($\mathbf{f}_{avg}=\sum_{i=1}^{n}AP(\mathbf{f}_i)$) while max pooling is performed over the feature map at the output of conv5. The contrastive feature $\mathbf{f}_{cont}$ is obtained by subtracting $\mathbf{f}_{max}$ from $\mathbf{f}_{avg}$($\mathbf{f}_{cont}=\frac{1}{n-1} (\mathbf{f}_{avg}-\mathbf{f}_{max})$), which denotes the discrepancy information between them. To reduce dimensionality, the bottleneck layer is employed to process  $\mathbf{f}_{cont} $ and $\mathbf{f}_{max}$ of the channel dimension $C$, and the corresponding reduced-dimensional features are denoted by $\mathbf{f}_{cont}'$ and $\mathbf{f}_{max}'$ of the channel dimension of $c$.  Then, the global contrastive feature $\mathbf{q}_0$ can be written as
\begin{eqnarray}
\mathbf{q}_0 = \mathbf{f}_{max}' + \mathcal{B}(\mathcal{C}(\mathbf{f}_{max}',\mathbf{f}_{cont}'))
\end{eqnarray}
where $\mathcal{C}$ means the concatenation of $\mathbf{f}_{cont} '$ and $ \mathbf{f}_{max}'$ to form a column vector with the channel dimension of $2c$, $\mathcal{B}$ represents the operation of bottleneck layer, with which the channel dimension is reduced to $c$ ($2c\rightarrow c$). 

\subsubsection{One-vs.-rest Relation Branch}
The fourth branch (v4) is the one-vs-rest relation branch.  In general, the part-level features often contain information about individual parts, which, however, does not reflect the relationship between them. The one-vs-rest relation branch makes it possible to associate each part-level with the corresponding rest parts. Similar to GCP, we first get 6 horizontal-grid features as $ (\mathbf{f}_1, \cdots, \mathbf{f}_6) $ and a 1536-dimensional vector through one-vs-rest relation module is computed as follows. Firstly,  AP is employed to get
\begin{eqnarray}
\mathbf{r}_i = \frac{1}{5}\sum_{j \neq i}\mathbf{f}_j. 
\end{eqnarray}
Then, both $ \mathbf{f}_i $ and $ \mathbf{r}_i $ are processed by the bottleneck layer for reducing the number of channels from C to c,  producing $ \mathbf{f}_i' $ and $ \mathbf{r}_i' $. With relation network, a local relational feature $\mathbf{q}_i$ can be computed as
\begin{eqnarray}
\mathbf{q}_i = \mathbf{f}_i' + \mathcal{B}(\mathcal{C}(\mathbf{f}_i',\mathbf{r}_i')), \quad i=1,\cdots,6
\end{eqnarray}
where $ \mathbf{q}_i $ is a 256-dimensional vector.

\begin{table*}[ht]
	\begin{center}
		\begin{tabular}{l|c@{\hskip 5pt}c@{\hskip 5pt}|c@{\hskip 5pt}c@{\hskip 5pt}|c@{\hskip 5pt}c@{\hskip 5pt}|c@{\hskip 5pt}c@{\hskip 5pt}}
			\toprule[1.5pt]
			\multirow{2}{*}{Global Features} &	\multicolumn{2}{c|}{CUHK03-Labeled} & \multicolumn{2}{c|}{CUHK03-Detected} & \multicolumn{2}{c|}{Market1501}	 &	\multicolumn{2}{c}{DukeMTMC}	\\
			\cline{2-9}
			&	mAP 	&	rank-1 &	mAP 	&	rank-1  &	mAP 	&	rank-1	&	mAP 	&	rank-1	\\
			\hline\hline
			HA-CNN\cite{Li2018HACNN} &41.0&44.4&38.6&41.7&75.7&95.6&63.8&80.5\\
			
			PCB\cite{Sun2018PCB} &-&-&54.2&61.3&77.3&92.4&65.3&81.9 \\
			
			AlignedReID\cite{Luo2019AlignedReID} &-&-&59.6&61.5&79.1&91.8&69.7&82.1 \\		
			
			PCB+RPP\cite{Sun2018PCB} &-&-&57.5&63.7&81.0&93.1&68.5&82.9 \\
			
			HPM\cite{Yang2018HPM} &-&-&57.5&63.9&82.7&94.2&74.3&86.6 \\
			
			BagOfTricks\cite{Luo2019bagoftricks} &-&-&-&-&85.9&94.5&76.4&86.4 \\
			
			OSNet\cite{Zhou2019OSNet} &-&-&67.8&72.3&84.9&94.8&73.5&88.6 \\
			
			MGN\cite{Wang2018MGN} &67.4&68.0&66.0&66.8&86.9&95.7&78.4&88.7 \\
			
			ABD\cite{Chen2019ABD} &-&-&-&-&88.28&95.6&78.59&89.0 \\
			
			GCP\cite{Park2019GCP} &75.6&77.9&69.6&74.4&88.9&95.2&78.6&89.7 \\						
			
			BDB\cite{Dai2019BDB} &76.7&79.4&73.5&76.4&86.7&95.3&76.0&89.0 \\
			
			SONA\cite{Xia2019SONA} &79.23&81.85&76.35&79.10&88.67&\bf{95.68}&78.05&89.25 \\
			
			\hline
			Ours  &\bf{84.0}&\bf{87.1}&\bf{80.5}&\bf{84.3}&\bf{89.5}&95.6&\bf{81.2}&\bf{91.4} \\
			\hline
		\end{tabular}
	\end{center}
	\caption{Comparison with the state-of-the-art methods for Market1501, DukeMTMC-reID and CUHK03 dataset in person reID}
	\label{tab1}
\end{table*}

\begin{table*}[ht]
	\begin{center}
		\begin{tabular}{l|c@{\hskip 5pt}c@{\hskip 5pt}|c@{\hskip 5pt}c@{\hskip 5pt}|c@{\hskip 5pt}c@{\hskip 5pt}|c@{\hskip 5pt}c@{\hskip 5pt}}
			\toprule[1.5pt]
			\multirow{2}{*}{Global Features} &	\multicolumn{2}{c|}{CUHK03-Labeled} & \multicolumn{2}{c|}{CUHK03-Detected} &\multicolumn{2}{c|}{Market1501}	 &	\multicolumn{2}{c}{DukeMTMC} 			\\
			\cline{2-9}
			&	mAP 	&	rank-1 &	mAP 	&	rank-1  &	mAP 	&	rank-1	&	mAP 	&	rank-1	\\
			\hline\hline
			local-global &79.7 &83.4 &75.4 &78.2&88.2&95.5&81.2&91.1 \\
			\hline
			local-global-OvR &83.0 &85.7 &79.2 &82.1&88.8&95.1&\bf{81.3}&90.8 \\
			\hline
			local-global-gcp &81.2 &83.9 &78.2 &81.9&89.5&95.5&80.7&90.3 \\
			\hline
			local-global-gcp-OvR &\bf{84.0} &\bf{87.1} &\bf{80.5} &\bf{84.3}&\bf{89.5}&\bf{95.6}&81.2&\bf{91.4} \\
			\hline
		\end{tabular}
	\end{center}
	\caption{The effects of various combinations of branches}
	\label{tab2}
\end{table*}

\begin{table*}[ht]
	\begin{center}
		\begin{tabular}{l|c@{\hskip 5pt}c@{\hskip 5pt}|c@{\hskip 5pt}c@{\hskip 5pt}|c@{\hskip 5pt}c@{\hskip 5pt}|c@{\hskip 5pt}c@{\hskip 5pt}}
			\toprule[1.5pt]
			\multirow{2}{*}{Global Features} &	\multicolumn{2}{c|}{CUHK03-Labeled} & \multicolumn{2}{c|}{CUHK03-Detected} &\multicolumn{2}{c|}{Market1501}	 &	\multicolumn{2}{c}{DukeMTMC} 		\\
			\cline{2-9}
			&	mAP 	&	rank-1 &	mAP 	&	rank-1  &	mAP 	&	rank-1	&	mAP 	&	rank-1	\\
			\hline\hline
			w/o-GeM &83.3 &86.5 &80.1 &83.6&\bf{89.8}&\bf{95.7}&\bf{81.3}&91.2 \\
			\hline
			w-GeM &\bf{84.0} &\bf{87.1} &\bf{80.5} &\bf{84.3}&89.5&95.6&81.2&\bf{91.4}\\
			\hline
		\end{tabular}
	\end{center}
	\caption{The impact of GeM on final performance, 'w/o-GeM' means without GeM, 'w-GeM' means with GeM}
	\label{tab3}
\end{table*}

\begin{table*}[ht]
	\begin{center}
		\begin{tabular}{l|c@{\hskip 5pt}c@{\hskip 5pt}|c@{\hskip 5pt}c@{\hskip 5pt}|c@{\hskip 5pt}c@{\hskip 5pt}|c@{\hskip 5pt}c@{\hskip 5pt}}
			\toprule[1.5pt]
			\multirow{2}{*}{Global Features} & \multicolumn{2}{c|}{CUHK03-Labeled} & \multicolumn{2}{c|}{CUHK03-Detected} &	\multicolumn{2}{c|}{Market1501}	 &	\multicolumn{2}{c}{DukeMTMC} 			\\
			\cline{2-9}
			&	mAP 	&	rank-1 &	mAP 	&	rank-1  &	mAP 	&	rank-1	&	mAP 	&	rank-1	\\
			\hline\hline
			f6+f4+f2 &82.6 &84.8 &79.0 &82.7 &89.5&\bf{95.7}&\bf{81.8}&\bf{91.7} \\
			\hline
			f6  &\bf{84.0} &\bf{87.1} &\bf{80.5} &\bf{84.3} &\bf{89.5}&95.6&81.2&91.4\\
			\hline
		\end{tabular}
	\end{center}
	\caption{Comparison between $\mathbf{q}^{h6}$ and $\mathcal{C}(\mathbf{q}^{h6},\mathbf{q}^{h4},\mathbf{q}^{h2})$}
	\label{tab4}
\end{table*}

\begin{table*}[ht]
	\begin{center}
		\begin{tabular}{l|c@{\hskip 5pt}c@{\hskip 5pt}|c@{\hskip 5pt}c@{\hskip 5pt}|c@{\hskip 5pt}c@{\hskip 5pt}|c@{\hskip 5pt}c@{\hskip 5pt}}
			\toprule[1.5pt]
			\multirow{2}{*}{Global Features} & \multicolumn{2}{c|}{CUHK03-Labeled} & \multicolumn{2}{c|}{CUHK03-Detected} &	\multicolumn{2}{c|}{Market1501}	 &	\multicolumn{2}{c}{DukeMTMC}		\\
			\cline{2-9}
			&	mAP 	&	rank-1 &	mAP 	&	rank-1  &	mAP 	&	rank-1	&	mAP 	&	rank-1	\\
			\hline\hline
			BC-OSNet &84.0 &87.1 &80.5 &84.3 &\bf{89.5}&\bf{95.6}&\bf{81.2}&\bf{91.4} \\
			\hline
			+GCDropout+BDB &\bf{84.2} &\bf{87.3} &\bf{81.6} &\bf{85.3} &89.2&95.3&80.7&91.0 \\
			\hline
		\end{tabular}
	\end{center}
	\caption{The impact of GCDropout and BDB on final performance}
	\label{tab5}
\end{table*}

\subsection{Loss Functions}
To train our model, the final total loss is the sum of the loss functions of each branch, including a single ID loss(softmax loss), a soft margin triplet loss\cite{Hermans2017tripletloss} and a center loss\cite{Wen2016centerloss}.
\begin{eqnarray}
L_{sum} = \lambda_1 L_{id} + \lambda_2 L_{triplet} + \lambda_3 L_{center}
\end{eqnarray}
where $\lambda_1,\lambda_2,\lambda_3$ are weighting factors. The single ID loss is shown in equation\ref{E2}. Randomly extract each small batch of $P$ identities and $K$ instances, soft margin triplet loss is defined as:
\begin{eqnarray}
%\begin{split}
\nonumber
L_{triplet}=\sum_{i=1}^{P} \sum_{a=1}^{K}[\alpha+ \overbrace{\max _{p=1 \ldots K}\left\|x_{a}^{(i)}-x_{p}^{(i)}\right\|_{2}}^{\text{hardest positive}}  \\  
-\underbrace{\min _{n=1 \ldots K \atop j=1 \ldots P , j \neq i}\left\|x_{a}^{(i)}-x_{n}^{(i)}\right\|_{2}}_{ \text{ hardest negative }} ]_{+}
%\end{split}
\end{eqnarray}
where $ x_a^{(i)}, x_p^{(i)}, x_n^{(i)} $ are features extracted from anchor, positive sample and negative sample, respectively, and $ \alpha $ is the edge hyperparameter. In order to improve the discriminative capability of features, the center loss is used as
\begin{eqnarray}
\mathcal{L}_C = \frac{1}{2}\sum_{i=1}^{m} \left\| \mathbf{x}_i - \mathbf{c}_{y_i} \right\|_2^2
\end{eqnarray}
where $ \mathbf{c}_{y_i} \in \mathbf{R}^d $ is the class center of the depth feature of class $y_i $.

%------------------------------------------------------------------------
\section{Experiments}
Extensive experiments were conducted on three person Re-ID dataset, including Market1501, DukeMTMC-reID and CUHK03.

\subsection{Datasets}

The Market1501 dataset is composed of 32668 pedestrian images taken by 6 cameras, a total of 1501 categories. The subset of train contains 12936 images of 751 identities, the subset of query contains 3368 images of 750 identities and the subset of gallery contains 15013 images of 751 identities. 

The DukeMTMC-reID dataset was captured by 8 cameras, including 16522 training images of 702 identities, 2228 query images of 702 identities and 17661 gallery images of 1110 identities. 

The CUHK03 dataset is divided into CUHK03\_labeled and CUHK03\_detected according to different annotation methods, of which there are 14096 images and 14097 images, respectively, which were captured by two cameras. For training subset, 7368 images for CUHK03\_label and 7365 images for CUHK03\_detected. 1400 images for both in query subset. For gallery subset, 5328 images for CUHK03 \_label and 5332 images for CUHK03\_detected.

\subsection{Implementation Details}

The input image is of size $256\times 128$ for both training and testing. The data augmentation methods include random flipping and random erasing\cite{Zhong2017random-rease}. The optimizer is Adam\cite{Kingma2014adam} with momentum of 0.9 and weight decay of 5e-04. During training, the batch size is set to 64 and the number of epoches is 160. Each batch contains 16 identities and each identity has 4 images. The warm up stragey is used during training, where the initial learning rate is 3.5e-4, after the 60th epoch, the learning rate is changed to 3.5e-05, and after the 130th epoch, the learning rate is changed to 3e-06. All networks are trained end-to-end using PyTorch. Training our model takes about sixteen, eighteen and eight hours with a singal NVIDIA Tesla P100 GPU for the Market1501, DukeMTMC-reID, and CUHK03 datasets, respectively.

\subsection{Comparison with State-of-the-art}

The comparison between BC-OSNet and state-of-the-art methods is shown in Table\ref{tab1} for the all three datasets. Note that none of the results below use re-ranking\cite{Zhong2017reranking} or multi-query fusion techniques. All recent methods for comparison include HA-CNN\cite{Li2018HACNN}, AlignedReID\cite{Luo2019AlignedReID}, PCB\cite{Sun2018PCB}, HPM\cite{Yang2018HPM}, MGN\cite{Wang2018MGN}, GCP\cite{Park2019GCP}, ABD\cite{Chen2019ABD}, BDB\cite{Dai2019BDB}, SONA\cite{Xia2019SONA}, OSNet\cite{Zhou2019OSNet}, et al. Clearly, BC-OSNet performs very competitively.

\subsection{Ablation Studies}

We conducted a large number of comparative experiments on Market-1501, DukeMTMC-reID and CUHK03 datasets to study the effectiveness of each branch, module and hyperparameters.

\subsubsection{Benefit of GCP and one-vs-rest relation Branches}

The proposed BC-OSNet comprehensively takes these four components into consideration. As shown in Table 2, competitive results are obtained on the three data sets, especially on the CUHK03 dataset. It can be seen that the four branches complement each other and can extract various features and manifest them in the experiment.

\subsubsection{Benefit of GeM}
GeM can be considered as a generalized version of GAP and GMP. Table\ref{tab3} shows that mAP and rank-1 have slightly increased on both CUHK03\_label and CUHK03\_detect. This means that it could learn a better version between GAP and GMP with the learnable parameter $p_k$ inherent in GeM.

\subsubsection{$\mathbf{q}^{h6} vs. \mathcal{C}(\mathbf{q}^{h6},\mathbf{q}^{h4},\mathbf{q}^{h2})$}
Let $\mathbf{q}^{h6} \triangleq \mathcal{C}(\mathbf{f}_0,\cdots,\mathbf{f}_6)$. We consider to use $\mathbf{q}^{h2}$ and $\mathbf{q}^{h4}$ that splits the initial feature map into two and four horizontal regions, respectively.  Accordingly, the concatenation of $\mathbf{q}^{h2}$, $\mathbf{q}^{h4}$ and $\mathbf{q}^{h6}$, namely, $\mathcal{C}(\mathbf{q}^{h2},\mathbf{q}^{h4},\mathbf{q}^{h6})$ is employed for final feature representation.  Note that $\mathbf{q}^{h2}$, $\mathbf{q}^{h4}$ and $\mathbf{q}^{h6}$ contain different local relational features, and thus have different global contrastive features. In\cite{Park2019GCP}, it was shown that the use of $ \mathcal{C} (\mathbf{q}^{h6}, \mathbf{q}^{h4}, \mathbf{q}^{h2}) $ could be better than the use of $\mathbf{q}^{h6}$. We, however, report the totally-different results as shown in Table\ref{tab4}. This may be the use of different backbone networks. 

\subsubsection{Benefit of GCDropout and BDB}
Ordinary dropout can effectively prevent overfitting. The dropout variables of Gaussian Continuous Dropout(GCDropout) are subject to a continuous distribution rather than the discrete distribution, moreover, the units in the network obey Gaussian distribution. GCDropout can effectively prevent the co-adaptation of feature detectors in deep neural networks and achieve a good balance between the diversity and independence of subnetworks. Batch DropBlock(BDB) force the network to learn some detailed features in the remaining area, which can complement the local branch. It can be seen from Table\ref{tab5} that the performance has been better improved after using those methods.

\section{Conclusion}
In this paper, we propose a branch-cooperative network for person Re-identification. Based on the OSNet baseline, we propose to a 4-branch architecture, with global, local, connection and contrast features, for obtaining more diverse and resolution features. In addition, various tricks have been incorporated into BC-OSNet, including GeM, Gaussian Continuous Dropout, et al. The ablation analysis clearly demonstrates the cooperation of four branches for boosting the final performance.  


\end{document}